\def\year{2020}\relax
\documentclass[letterpaper]{article} 
\usepackage{aaai20}  
\usepackage{times}  
\usepackage{helvet} 
\usepackage{courier}  
\usepackage[hyphens]{url}  
\usepackage{multirow}
\usepackage{multicol}
\usepackage{subfig}
\usepackage{graphicx} 
\usepackage{graphicx}  
\title{Cooperative Reasoning on Knowledge Graph and Corpus: A Multi-agent Reinforcement Learning Approach }
\begin{document}
\author{\\Yunan Zhang,\textsuperscript{1}
Xiang Cheng,\textsuperscript{1}
Heting Gao, \textsuperscript{1}
Chengxiang Zhai\textsuperscript{1}\\
\textsuperscript{1}{University of Illinois at Urbana-Champaign}\\
\{yunanz2, xiangc2, hgao17,czhai\}@illinois.com}

\maketitle

\begin{abstract}
Knowledge-graph-based reasoning has drawn a lot of attention due to its interpretability. However, previous methods suffer from the incompleteness of the knowledge graph, namely the interested link or entity that can be missing in the knowledge graph(explicit missing). Also, most previous models assume the distance between the target and source entity is short, which is not true on a real-world KG like Freebase(implicit missing). The sensitivity to the incompleteness of KG and the incapability to capture the long-distance link between entities have limited the performance of these models on large KG. In this paper, we propose a model that leverages the text corpus to cure such limitations, either the explicit or implicit missing links. We model the question answering on KG as a cooperative task between two agents, a knowledge graph reasoning agent and an information extraction agent. Each agent learns its skill to complete its own task, hopping on KG or select knowledge from the corpus, via maximizing the reward for correctly answering the question. The reasoning agent decides how to find an equivalent path for the given entity and relation. The extraction agent provide shortcut for long-distance target entity or provide missing relations for explicit missing links with messages from the reasoning agent. Through such cooperative reward design, our model can augment the incomplete KG strategically while not introduce much unnecessary noise that could enlarge the search space and lower the performance.
\end{abstract}

\section{Introduction}
\begin{figure}[t]
  \includegraphics[width=0.46\textwidth]{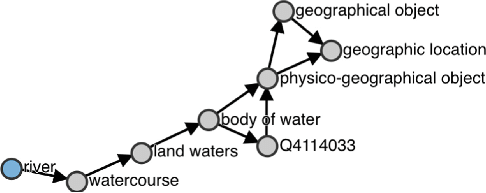}
  \caption{An example of implicit incompleteness on WikiData. To get the attribute from the the class river, we need to do 6 hops of jumping., river$ \rightarrow $watercourse$ \rightarrow $ land waters $ \rightarrow $body  of water $ \rightarrow $ physic-geographical object $ \rightarrow $ geographic location $ \rightarrow $ specific attribute.}
 
\end{figure}
Recently, knowledge graph reasoning has made a lot of progress. Previous methods can be categorized into three directions, namely, path-ranking based, embedding based and reinforcement learning. However, all these methods share similar weaknesses. The incompleteness of KG. We categorized the incompleteness into two kinds. The first one is the explicit incompleteness. We define this condition as when there's no path between the target entity and the source entity or when the target entity itself is missing. The random walk based methods and RL based methods are extremely prone to this kind of incompleteness because they certainly can't sample the right path. For example, we are interested in finding Obama's grandfather. But if there's no path between Obama and Stanley Armour Dunham, either one hop path like 'grand father of' or multi-hop path like 'father of -> father of', or the Stanley Armour Dunham entity itself doesn't exist, walking based model would naturally fail.
The second one is the implicit incompleteness, which we refer to the condition when there's a path between the target entity and source entity, but the path is so long that it's hard for models to infer. For example, given the query, "Which river is in Urbana?". To answer the query in WikiData\cite{DBLP:conf/www/Vrandecic12}, we need to conduct six hops. However, in experiments, most existing models won't be able to infer an equivalent path that is longer than 4 steps for a give source entity and relation. This also explains why so far, most knowledge graph models can only work on relative small datasets but usually fail on real-world industry-scale datasets like Freebase\cite{DBLP:conf/www/2012c} and Wikidata where there are millions of entities and there's no guarantee on the assumed short distance between entities. One solution to such incompleteness is knowledge base population, which is the task of building or extending a knowledge base from text. However, this line of methods has its limitations. The enrichment is conduct on a static knowledge graph. It's hard to say whether such extension is optimal for a specific task, e.g., answering certain kinds of questions. This can be seen in two aspects. First, the specific knowledge can still be missing for a given sample because the KBP is not optimized dynamically for each question. The enriched graph can be larger than necessary while still missing interested triples. Second, such enrichment introduce a lot of noises that is much beyond the amount needed to answer a specific question, thus unnecessarily increase the search space of the model. \\
\\
To handle these limitations, we design a principled way to dynamically enrich the KG corresponding to the reasoning state of the model via cooperative multi-agent reinforcement learning\cite{DBLP:conf/icml/Tan93}. Our framework defines two types of agents that are expected to learn their skills, reasoning, and extraction, to jointly maximize the reward of the whole system. The reasoning agent decides which outgoing edge and the end side entity of the current entity it should hop to. This path-finding is similar to previous KG reasoning models. The extraction agent decides which extra triple to add to the KG given the current state of the system. The triple to add depends on the current query solving the state of the system, which could dynamically assist the reasoning agent with timely missing information on the KG. The two agents use a communication channel to share their observations and fuse the message with its local state to decide what action to take. In this way, the model dynamically decides when and how to enrich the KG to realize good performance and efficient operations.\\
\\
We explore credit assignment of the collaboration along several modes. (1)Only a final reward is given to both the agents upon the task completion. (2)A final reward the given to the reasoning agent if it reaches the correct target entity. But extraction agent only receive reward when the correct path include the proposed extracted triple, namely when the reasoning agent adopt the added triples. Though such reward design is actually not cooperative and there's no guarantee that the extraction policy would converge. (3)A reward shaping approach for the failed path. If it failed, we still assign a reward based on the distance between the target entity and end entity. (4) A game-theoretic setting, where the reasoner tries to minimize its number of hops to the target entity and the extractor tries to minimize the number of rejected proposals. Both agents are only rewarded when the correct answer is found.\\
\\
To conclude, our contributions are three-fold:
\begin{enumerate}
	\item Study the dynamic knowledge graph completion task.
	\item Propose a novel framework based on cooperative reinforcement learning for this task.
	\item Experiment and analysis shows the potential of this direction.
\end{enumerate}

\section{Related Work}
We now review related work in the directions of knowledge graph reasoning, multi-agent reinforcement learning, and fusion of KB and texts for question answering.
\subsection{Knowledge Graph Reasoning}
There have been many attempts in this area over the past few years. This task is motivated by the fact the though relation in the query $<e_s,r, e_t>$ can be missing in the KG, but there' an equivalent path that leads to $e_t$, e.g., grandfather relation is missing between$e_s$ and $e_t$ but there is two father of and an intermediate entity between them. In this case, the incompleteness is alleviate via finding equivalent path. They can be generally categorized into three parts. First, Path-Ranking Algorithm (PRA) method\cite{DBLP:conf/emnlp/LaoMC11,DBLP:conf/emnlp/GardnerTKM13,DBLP:conf/emnlp/GardnerTKM14,DBLP:conf/acl/WangC15}, which is based on random walk with restart to aggregate feature of neighborhood of nodes to make predictions. One issue with the method is combinatorially increasing space which slow down the inference and affect the accuracy. The second line of approach is embedding based methods. These works learn embedding of entities and relations using neural network or tensor factorization methods and use the embedding to predict how likely two entities are connected via certain relation\cite{DBLP:conf/emnlp/GuML15,DBLP:conf/aaai/NickelRP16,DBLP:conf/esws/SchlichtkrullKB18,DBLP:conf/aaai/DettmersMS018}. The third line is reinforcement learning based approaches\cite{DBLP:conf/aitest/WangWFCC19,DBLP:conf/iclr/DasDZVDKSM18,DBLP:conf/emnlp/LinSX18,DBLP:conf/naacl/GodinKM19,DBLP:conf/nips/ShenCHGG18}, which is the state of the art method in this task. It models the multi-hop path completion as a Markov decision process(MDP) and use RL to optimize the strategy of choosing the entity and relation pair to jump to given current entity and query.
\subsection{Multi-agent Reinforcement Learning}
Multi-agent systems have been explored broadly over the years. Relevant algorithms can be categorized via several features, (i) centralized or decentralized control\cite{DBLP:conf/icra/Pape90}, which mainly differs in whether there's an agent control all other agents, (ii) cooperative or competitive environment\cite{DBLP:conf/ecai/Galliers88}, which mainly differs in whether each agents shares a common target to maximize or then compete with each other to maximize one's own reward. Early attempts apply relevant algorithms to predator-prey game, grid-maze games and other syntactic environment. Recently, it has been used in visual navigation\cite{DBLP:conf/cvpr/JainWKRLFSK19,DBLP:conf/naacl/CelikyilmazBHC18}, object detection \cite{DBLP:conf/cvpr/KongXWH17}, dialogue\cite{DBLP:conf/emnlp/KotturMLB17}, and summarization. Recent progress in its application mainly focuses on the communication protocol, which demonstrates superior performance to non-communicating counterparts. In this work, we mainly focus on the benefits of a decentralized, fully observable, cooperative environment.
\subsection{Combination of KB and Texts for Question Answering}
There are several attempts for developing hybrid of text-based and KB-based QA systems\cite{DBLP:conf/acl/DasZRM17}, which stores text and KB information into the same structure based on the universal schema\cite{DBLP:conf/naacl/RiedelYMM13}. Such system are also used to augment language model\cite{DBLP:journals/corr/AhnCPB16}, natural language inference\cite{DBLP:conf/naacl/AnnervazCD18} and question answering\cite{DBLP:conf/emnlp/SunDZMSC18}. There are mainly two style fusion. One is early fusion, the other is late fusion\cite{DBLP:conf/smc/GunesP05}. Early fusion fuse the two sources of information before the inference, namely the it fuses the feature vector of different sources before doing inferences, while late fusion  first conduct inferences on models trained via different features and then combines the prediction of them.
Follow up work  
\section{Approach}
In this section, we introduce our model in details. Our framework jointly trains two agents, a KG reasoning agent on a KG \textbf{G} and a information extraction agent on a corpus \textbf{C}. Given a structured query $(e_s, r_q)$, where $e_s$ is the start entity and $r_q$ is the interested relation, our model outputs the $e_t$ which is the entity connected to $e_s$ via $r_q$, and the path to $e_t$ if $r_q$ doesn't exist in the given KG. The KG reasoning agent decides which outgoing edge/relation from the current entity it should go to and reach the new entity and repeat such process until it stops via self-loop or reaches the maximum step limit. The IE agent ranks the knowledge in a given corpus \textbf{C} and adds the highest ranking $(e_i, r_i, e_{ti})$ to \textbf{G}. We design four different reward mechanics for the agents. (1) Fully cooperative, where agents receives the same reward upon successful completion of the task, (2) Cooperative but the extraction's reward also depends on whether the reasoner's adopt its proposal, (3) Cooperative with reward shaping, which is same as (2) but we assign soft reward even if the agent doesn't get the golden answer, (4) game-theoretic, where two agents have a shared task of correctly answering question, but each agent wants to minimize their cost. For the reasoner, it's the number of hops it reaches the answer. For the extractor, it's the number of rejected proposals by the reasoner. 
\subsection{The KG Reasoning Agent}
In this section, we formally define the reasoning agent. A knowledge graph can be represented as $\mathcal{G = \{\mathcal{E,R} \}}$, where $\mathcal{E}$ is the set of entities and $\mathcal{R}$ is the set of relations. A KB fact, or triple, is stored as, ($e_1$, r, $e_2$) $\in \mathcal{G}$, where $e_1$ is the source entity, $e_2$ is the target entity, r is the relation connecting $e_1$ and $e_2$, which is a directed edge pointing from $e_1$ to $e_2$.\par
The reasoning agent tries to answer query $(e_s, r_q, ?)$ via path finding, where $e_s$ is the source entity and $r_q$ is the relation of interest. Answering here means find the entity that is connected to $e_s$ by $r_q$.
Find a path means finding a path to ? so that even $(e_s, r_q, ?) \notin G$ due to incompleteness, we still find the set of answers $E_o = \{e_o\}$.
\subsubsection{Action}
The action space can be written as:\par $A_t = \{(e_{t-1},\hat{r_t}, e_t)| (e_{t-1},\hat{r_t}, e_t)\in \mathcal{G}\}$,\\ \\namely choosing a triple whose starting from source entity $e_s$. We have a self-loop action for every node to indicate the answer.
$\hat{r_t},e_t$is outgoing edge of $e_s$ and the reached entity.
\subsubsection{State}
We represent state as  $st = (et,(e_s, r_q)) \in S$, where $e_t$ is the entity visited at step t and
$(e_s, r_q)$ are the source entity and relation in the original query. In experiment, we uses ConvE for their representation. One issue though, is when $r_q$ never appears in any path of the graph and in the training set of ConvE. We adopt a simple zero-shot learning approach to handle this situation.
\subsubsection{Transition}
Transition can be represented as a probability matrix:\par 
$P(s_{t+1} = s'| s_{t}=s, a_t = a)$.
\subsubsection{Reward}
We design four kinds of reward, which we will elaborate in section 3.4.2 and discuss their pros and cons in the experiment section.
\subsection{The Information Extraction Agent}
In this section, we formally define the IE agent. The IE agents learns how to rank the most useful knowledge from corpus to help the reasoning agent. Given the reasoning agent is at $e_t$ at timestep t, the IE agent will rank all the triple it extracted from corpus of form  $\mathcal{K} = \{(e_t, r, e_d) \cup (e_d, r^{-1}, e_t)\}$ and add the highest ranking triple to \textbf{G}. 
\subsubsection{Action}
The action space is all the extracted triples:\par $\mathcal{K} = \{(e_t, r, e_d) \cup (e_d, r^{-1}, e_t)\}$,\\ \\
namely choosing the highest ranked triple and to $e_s$ in \textbf{G}.
\subsubsection{State}
The extraction receives a message from the reasoning agent, which is its state representation. We then fuse the representation with the concatenation of all the extracted feasible triples. We want the state to aware of what the reasoning agent can get and can't get in its neighborhood, so we pass a message from reasoning agent to the IE agent.
\subsubsection{Transition}
Transition can be represented as a probability matrix:\par 
$P(s_{t+1} = s'| s_{t}=s, a_t = a)$. 
\subsubsection{Reward}
We also design four kinds of reward, which we will also elaborate in section 3.4.2 and discuss their pros and cons in the experiment section.
\subsection{Policy Learning}
Typical MARL usually encounter non-stationary environment which makes training process quite challenging due to difficulty in converge to a fixed optimal strategy for a moving target. In our setting,  we adopt the Forget Algorithm and Respond to Target Opponents Algorithm to handle this challenge.
\subsubsection{Handling non-stationary environment}
For the Forget algorithm, we use R-max\cite{DBLP:journals/jmlr/BrafmanT02} to handle the changing environment, which conduct a drift exploration to detect changes that happened in the opponent and continually revisit states that have not been visited recently. In this way, the model is guarantee to fit to the newly added triples by the IE agent. This handing is used for the first three reward designs.\par
For the  Respond to Target Opponents Algorithm, we use Minimax game\cite{DBLP:conf/icml/Littman94} specifically for the game-theoretic reward setting. 
\subsubsection{Reward Design}
We design 4 kinds of reward. The first is the most simple setting, namely both agents receive a +1 reward for correctly returning the answer.\par
The second reward setting does the same thing to the KG reasoning agent but only assigns a reward to the extraction agent when the reasoning agent adopt its proposal.\par
The third setting is similar to the second one, but we assign a soft reward to the agents when they stop at the wrong answer. The soft reward is given by the similarity between the output entity and the golden entity, which is measured by distance between their embeddings. \par
The fourth reward setting is a game theoretic one. It's similar to the third one but each agent has its own target. The reasoning agent wants to minimize the number of hops it takes to reach the correct the answer while the extraction agent wants to minimize the number of unused added triples.
\section{Discussion}
In this paper, we explore and discuss the potential of using MARL to effectively and efficiently solve the knowledge graph reasoning issue. We propose several possible framework that may or may not useful to solve some bottlenecks of this task. One of my concern is how to represent the state of the agent. As we mentioned earlier, the node embedding used for state is trained on the KG before update. However, every time we add triples to the KG, the corresponding embedding should also change via re-training it on the updated graph. Such pipeline would be computation heavy. But on the other hand, it's not easy to find a representation method that is better than node embedding to work well on large graphs.
\section{Acknowledgements}
This work is a manuscript of Yunan Zhang. The work is devoted to Makise Kurisu, despite its limited novelty. We thank Suhansanu Kumar for insightful discussions.

\bibliography{aaai20}
\bibliographystyle{aaai20}
\end{document}